\documentclass[11pt,a4paper]{article}
\usepackage[hyperref]{acl2021}
\usepackage{times}
\usepackage{latexsym}

\usepackage{times}
\usepackage{url}
\usepackage{algorithm}
\usepackage{multicol}
\usepackage{algorithmicx}
\usepackage{mathrsfs}
\usepackage{latexsym}
\usepackage{graphicx}
\usepackage{float}
\usepackage{amsmath, bm}
\usepackage{array}
\usepackage{algpseudocode}

\usepackage{microtype}

\usepackage{microtype}

\aclfinalcopy 

\title{LearnDA: Learnable Knowledge-Guided Data Augmentation for \\ Event Causality Identification}

\author{Xinyu Zuo$^{1,2}$, Pengfei Cao$^{1,2}$, Yubo Chen$^{1,2}$, Kang Liu$^{1,2}$, Jun Zhao$^{1,2}$, \\
\textbf{Weihua Peng}$^{3}$ \and \textbf{Yuguang Chen}$^{3}$
 \\
	$^1$National Laboratory of Pattern Recognition, Institute of Automation,
 CAS, Beijing, China\\
 $^2$School of Artificial Intelligence, University of Chinese Academy of Sciences, Beijing, China \\
 $^3$Beijing Baidu Netcom Science Technology Co., Ltd \\
	{\tt \{xinyu.zuo,pengfei.cao,yubo.chen,kliu,jzhao\}@nlpr.ia.ac.cn} \\
	{\tt \{pengweihua,chenyuguang\}@baidu.com}
	} 

\date{}

\begin{document}
\maketitle
	\begin{abstract}
		Modern models for event causality identification (ECI) are mainly based on supervised learning, which are prone to the data lacking problem. Unfortunately, the existing NLP-related augmentation methods cannot directly produce available data required for this task. To solve the data lacking problem, we introduce a new approach to augment training data for event causality identification, by iteratively generating new examples and classifying event causality in a dual learning framework. On the one hand, our approach is knowledge guided, which can leverage existing knowledge bases to generate well-formed new sentences. On the other hand, our approach employs a dual mechanism, which is a learnable augmentation framework, and can interactively adjust the generation process to generate task-related sentences. Experimental results on two benchmarks EventStoryLine and Causal-TimeBank show that 1) our method can augment suitable task-related training data for ECI; 2) our method outperforms previous methods on EventStoryLine and Causal-TimeBank (+2.5 and +2.1 points on F1 value respectively).
		
	\end{abstract}
	
	\section{Introduction}
	\label{intro}
\begin{figure}[t]
		\centering
		\includegraphics*[clip=true,width=0.48\textwidth,height=0.10\textheight]{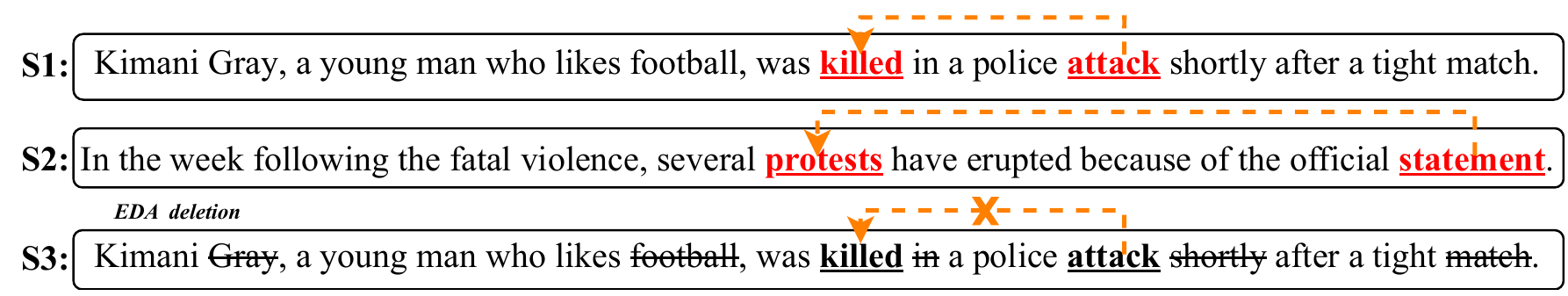}
		\caption{S1 and S2 are \emph{causal sentences} that contain \emph{causal events}. S3 is produced by EDA based on S1. The dotted line indicates the causal relation.} \label{fig1}
	\end{figure}

	Event causality identification (ECI) aims to identify causal relations between events in texts, which can provide crucial clues for NLP tasks, such as logical reasoning and question answering \cite{girju2003automatic,oh2013question,oh2017multi}. This task is usually modeled as a classification problem, i.e. determining whether there is a causal relation between two events in a sentence. For example in Figure \ref{fig1}, an ECI system should identify two causal relations in two sentences: (1) \textbf{attack} $ \stackrel{cause}{\longrightarrow}$ \textbf{killed} in S1; (2) \textbf{statement} $\stackrel{cause}{\longrightarrow}$ \textbf{protests} in S2. 
	
	Most existing methods for ECI heavily rely on annotated training data \cite{mirza2016catena,riaz2014recognizing,hashimoto2014toward,hu2017inferring,gao-etal-2019-modeling}. However, existing datasets are relatively small, which impede the training of the high-performance event causality reasoning model. According to our statistics, the largest widely used dataset EventStoryLine Corpus \cite{caselli2017event} only contains 258 documents, 4316 sentences, and 1770 causal event pairs. Therefore, data lacking is an essential problem that urgently needs to be addressed for ECI.
	
	Up to now, data augmentation is one of the most effective methods to solve the data lacking problem. However, most of the NLP-related augmentation methods are a task-independent framework that produces new data at one time \cite{Zhang2015CharacterlevelCN,Guo2019AugmentingDW,Xie2019UnsupervisedDA}. In these frameworks, data augmentation and target task are modeled independently. This often leads to a lack of task-related characteristics in the generated data, such as task-related linguistic expression and knowledge. For example, easy data augmentation (EDA) \cite{wei-zou-2019-eda} is the most representative method that relies on lexical substitution, deletion, swapping, and insertion to produce new data. 
	However, solely relying on such word operations often generates new data that dissatisfies task-related qualities.
	As shown in Figure \ref{fig1}, S3 is produced by EDA, it lacks a linguistic expression that expresses the causal semantics between \emph{kill} and \emph{attack}. Therefore, how to interactively model data augmentation and target task to generate new data with task-related characteristics is a challenging problem on ECI.
	
	
	Specific to ECI, we argue that an ideal task-related generated causal sentence needs to possess two characteristics as follows. (1) 
	The two events in the causal sentence need to have a causal relation. We call such property as \textbf{\emph{Causality}}. For example, there is usually a causal relation between an \emph{attack} event and a \emph{kill} event, while nearly no causal relation between an \emph{attack} event and a \emph{born} event. (2) The linguistic expressions of the causal sentence need to be well-formed to express the causal semantic of events. We call such property as \textbf{\emph{Well-formedness}}, which consists of a) canonical sentence grammar, b) event-related entities with semantic roles (e.g. the \emph{attack} was carried out by a \emph{police} in S1), and c) cohesive words that express complete causal semantics (e.g. \emph{in a} and other words except for events and entities in S1).
	
	
	To this end, we propose a learnable data augmentation framework for ECI, dubbed as \textbf{Learn}able Knowledge-Guided \textbf{D}ata \textbf{A}ugmentation (LearnDA). This framework regards sentence-to-relation mapping (\emph{the target task}, ECI) and relation-to-sentence mapping (\emph{the augmentation task}, sentence generation) as dual tasks and models the mutual relation between them via dual learning. Specifically, LearnDA can use the duality to generate task-related new sentences learning from identification and makes it more accurate to understand the causal semantic learning from generation. On the one hand, LearnDA is knowledge guided. It introduces diverse causal event pairs from KBs to initialize the dual generation which could ensure the \textbf{\emph{causality}} of generated causal sentences. For example, the knowledge of \emph{judgment} $\stackrel{cause}{\longrightarrow}$ \emph{demonstration} from KBs can be used to construct a novel causal sentence, which is also helpful to understand the causal semantic of \emph{statement} $\stackrel{cause}{\longrightarrow}$ \emph{protests}. On the other hand, LearnDA is learnable. It employs a constrained generative architecture to generate \textbf{\emph{well-formed}} linguistic expressions via iteratively learning in the dual interaction, which expresses the causal semantic between given events. Methodologically, it gradually fills the remaining missing cohesive words of the complete sentences under the constraint of given events and related entities.
	
	In experiments, we evaluate our model on two benchmarks. We first concern the standard evaluation and show that our model achieves the state-of-the-art performance on ECI. Then we estimate the main components of LearnDA. Finally, our learnable augmentation framework demonstrates definite advantages over other augmentation methods in generating task-related data for ECI.
	
	In summary, the contributions as follows:
	\begin{itemize}
		\item We propose a new learnable data augmentation framework to solve the data lacking problem of ECI. Our framework can leverage the duality between identification and generation via dual learning which can learn to generate task-related sentences for ECI.
		
		\item Our framework is knowledge guided and learnable. Specifically, we introduce causal event pairs from KBs to initialize the dual generation, which could ensure the causality of generated causal sentences. We also employ a constrained generative architecture to gradually generate well-formed causal linguistic expressions of generated causal sentences via iteratively learning in the dual interaction.
		
		\item Experimental results on two benchmarks show that our model achieves the best performance on ECI. Moreover, it also shows definite advantages over previous data augmentation methods.
	\end{itemize}
	
	
	\section{Related Work}
	To date, many researches attempt to identify the causality with linguistic patterns or statistical features. For example, some methods rely on syntactic and lexical features \cite{riaz2013toward,riaz2014recognizing}. Some focus on explicit causal textual patterns \cite{hashimoto2014toward,riaz2014depth,riaz2010another,do2011minimally,hidey-mckeown-2016-identifying}. And some others pay attention on statistical causal association and cues \cite{beamer2009using,hu2017inference,hu2017inferring}. 
	
	Recently, more attention is paid to the causality between events. \citet{Mirza2014AnAO} annotated Causal-TimeBank of event-causal relations based on the TempEval-3 corpus. \citet{mirza2014annotating}, \citet{mirza2016catena} extracted event-causal relation with a rule-based multi-sieve approach and improved the performance incorporating with event temporal relation. \citet{mostafazadeh2016caters} annotated both temporal and causal relations in 320 short stories. \citet{caselli2017event} annotated the EventStoryLine Corpus for event causality identification. \citet{dunietz-etal-2017-corpus} presented BECauSE 2.0, a new version of the BECauSE corpus \cite{dunietz-etal-2015-annotating} of causal relation and other seven relations. \citet{gao-etal-2019-modeling} modeled document-level structures to identify causality. \citet{ijcai2020-499} identified event causality with the mention masking generalization.
    
    Unlike computer vision, the augmentation of text data in NLP is pretty rare \cite{chaudhary2020nlpaugment}. \citet{zuo-etal-2020-knowdis} solved the data lacking problem of ECI with the distantly supervised labeled training data. However, including the distant supervision, most of the existing data augmentation methods for NLP tasks are task-independent frameworks (Related work of data augmentation and dual learning are detailed in Appendix B). Inspired by some generative methods which try to generate additional training data while preserving the class label \cite{AnabyTavor2019NotED,yang-etal-2019-exploring-pre,Papanikolaou2020DAREDA}, we introduce a new learnable framework for augmenting task-related training data for ECI via dual learning enhanced with external knowledge.

	\begin{figure}[t] 
		\centering
		\includegraphics*[clip=true,width=0.41\textwidth,height=0.19\textheight]{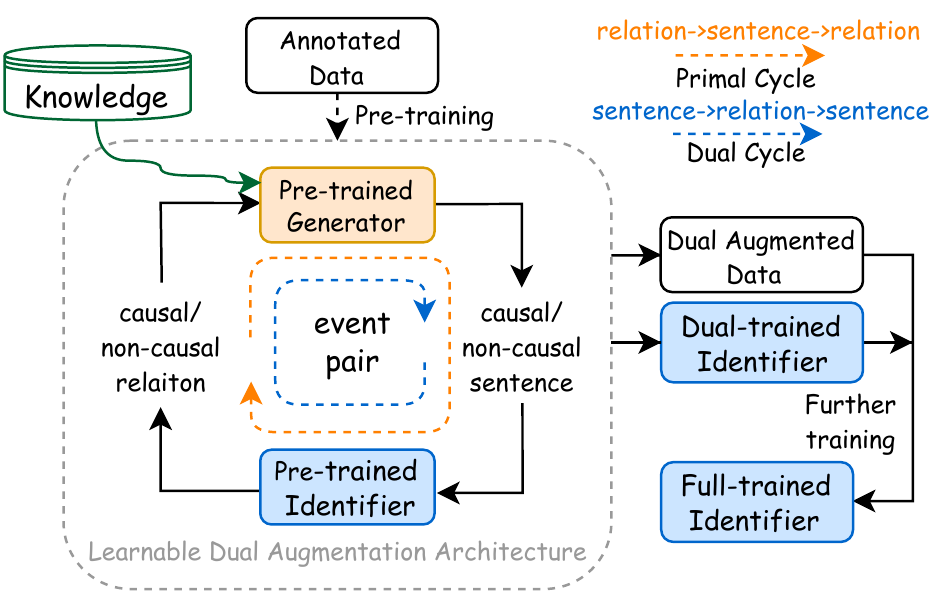}
		\caption{Overview of the learnable knowledge-guided dual data augmentation for ECI.} \label{fig2}
	\end{figure}
	
	\section{Methodology}
	As shown in Figure \ref{fig2}, LearnDA jointly models a knowledge guided sentence generator (input: \emph{event pair and its causal/non-causal relation}, output: \emph{causal/non-causal sentence})  and an event causality identifier (input: \emph{event pair and its sentence}, output: \emph{causal/non-causal relation}) with dual learning. LearnDA iteratively optimizes identifier and generator to generate task-related training data, and then utilize new data to further train the identifier. Therefore, we first present the main idea of dual learning, which is the architecture of learnable dual augmentation, including the states, actions, policies, and rewards. Then, we briefly introduce the knowledge guided sentence generator, especially the processes of knowledge guiding and constrained sentence generation. Finally, we describe the event causality identifier and training processes of LearnDA.
	
	\subsection{Architecture of Learnable Dual Augmentation}
	\label{sec:LDAN}
	\begin{figure}[t] 
    	\centering
		\includegraphics*[clip=true,width=0.39\textwidth,height=0.19\textheight]{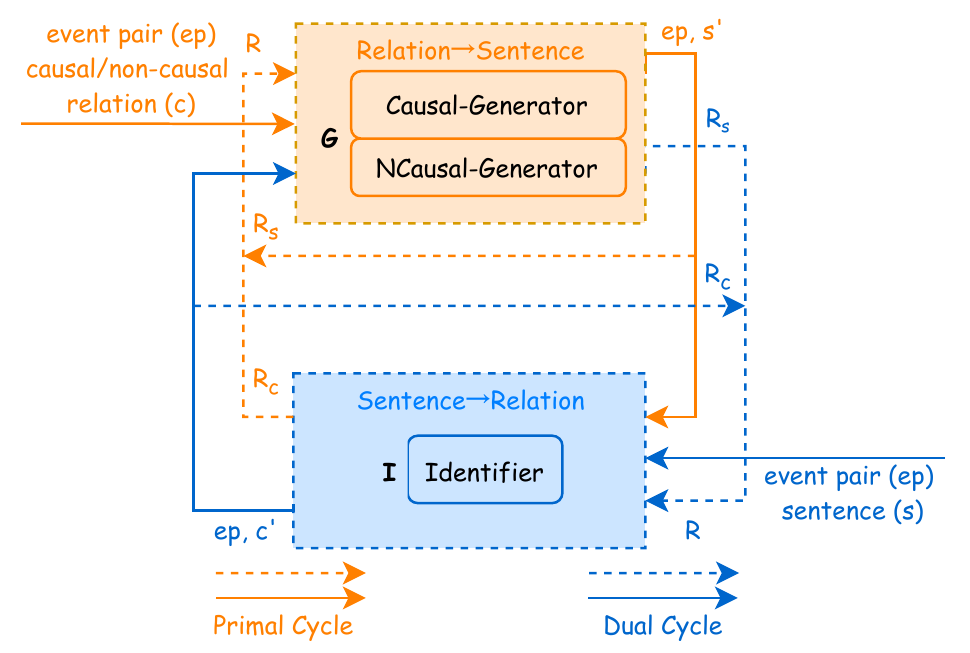}
		\caption{The architecture of learnable dual augmentation. \emph{Causal} and \emph{NCausal} represent the causal and non-causal sentence generator respectively. Red parts are the process of \emph{$<$event pair, relation$>$ $\rightarrow$ sentence $\rightarrow$ relation} (primal cycle), while blue parts are the process of \emph{$<$event pair, sentence$>$ $\rightarrow$ relation $\rightarrow$ sentence} (dual cycle). Solid and dashed lines denote the main process and reward feedback direction respectively.} \label{fig3}
	\end{figure}
	
	The architecture of learnable dual augmentation is shown in Figure \ref{fig3}. Specifically, \emph{I} denotes the event causality identifier, and \emph{G} denotes the sentence generator which consists of two independent generators. They produce causal and non-causal sentences on the relation $c$ of input event pair $ep$. 
	
	Generally, \emph{G} generates a sentence $s'$ which expresses the causal or non-causal relation $c$ of the input event pair $ep$. Then it receives the reward $R$ that consists of a semantic alignment reward $R_s$ from itself and a causality reward $R_c$ from \emph{I} (primal cycle). Similarly, \emph{I} identifies the causal or non-causal relation $c'$ of the input event pair $ep$ with its sentence $s$. Then it receives the reward $R$ consists of a causality reward $R_c$ from itself and a semantic alignment reward $R_s$ from \emph{G} (dual cycle). 
	
	\emph{I} and \emph{G} are optimized interactively with dual reinforcement learning. Specifically, for \emph{G}, an action is the generation from relation to sentence, a state is denoted by the representation of input event pair and its relation, a policy is defined by the parameters of generator. For \emph{I}, an action is the identification from sentence to relation, a state is denoted by the representation of input event pair and its sentence, a policy is defined by the parameters of identifier. Inspired by \citet{shen-feng-2020-cdl}, we utilize a probability distribution over actions given states to represent the policys, i.e., the probability distribution of the generation of \emph{G} and identification of \emph{I}. As aforementioned, we introduce two rewards, causality ($R_c$) and semantic alignment ($R_s$) rewards, which encourage \emph{G} to generate task-related sentences with the feedback from identifier, while further optimize \emph{I} with the feedback from generator. Definitions are as following:
	
	\paragraph{Causality Reward ($R_c$)} If the relation of input event pair can be clearly expressed by the generated sentence, it will be easier to be understood by identifier. Therefore, we use the causal relation classification accuracy as the causality reward to evaluate the causality of generated sentences, while tune and optimize the identifier itself:
	\begin{equation}\footnotesize 
	\setlength{\abovedisplayskip}{3pt} 
	\setlength{\belowdisplayskip}{3pt}
	R_{c}(ep, s) = 
	\begin{cases}
	p(c'|s;\theta_{I})& \text{\emph{Correct classification}}\\
	-p(c'|s;\theta_{I})& \text{\emph{Otherwise},}
	\end{cases} 
	\end{equation}
	where $\theta_{I}$ is the parameter of \emph{I}, $p(c'|s;\theta_{I})$ denotes the probability of relation classification, $s$ denotes the input sentence and $c'$ is the classified relation.
	
	\paragraph{Semantic Alignment Reward ($R_s$)} We hope that the semantic of the generated sentence can be consistent with the relation of the input event pair. Additionally, if the relation of the input event pair can be more accurately classified, the semantic of the new generated sentence can be considered more consistent with it. Therefore, we measure the semantic alignment by means of the probability of constructing a sentence with similar semantic to the input relation, and the reward is:
	\begin{equation}\footnotesize
	\setlength{\abovedisplayskip}{6pt} 
	\setlength{\belowdisplayskip}{6pt}
	R_{s}(ep, c) = p(s'|c;\theta_{G}) = \frac{1}{|T_s|} \sum_{t \in T_s} p(t|c;\theta_{G}),
	\end{equation}
	where $\theta_{G}$ is the parameter of \emph{G}, $c$ is the input relation, $t$ is one of the generated tokens $T_s$ of the generated sentence $s'$, and $p(t|c;\theta_{G})$ is the generated probability of $t$. Specifically, there are two independent \emph{G} with different $\theta_{G}$. In detail, $\theta_{G}^{c}$ is employed to generated causal sentence when the input $c$ is causal relation, and non-causal sentence is generated via $\theta_{G}^{nc}$ when $c$ is non-causal relation.
	
	\subsection{Knowledge Guided Sentence Generator}

\begin{figure}[t] 
		\centering
		\includegraphics*[clip=true,width=0.42\textwidth,height=0.17\textheight]{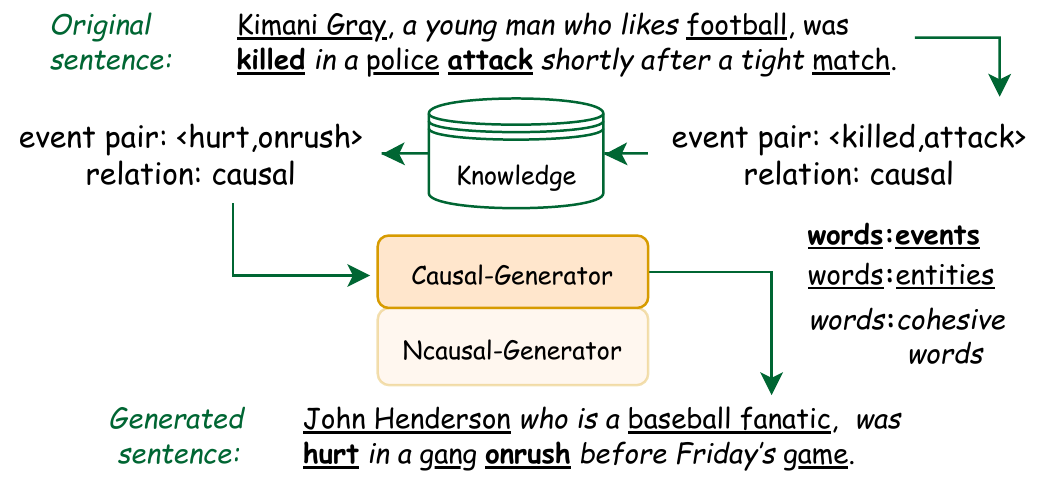}
		\caption{Flow diagram of the knowledge guided sentence generator (KSG). We take causal sentence generation via lexical knowledge expanding as an example.} \label{fig4}
	\end{figure}

	As shown in Figure \ref{fig4}, knowledge guided sentence generator (KSG) first introduces diverse causal and non-causal event pairs from KBs for \emph{causality}. Then, given an event pair and its causal or non-causal relation, it employs a constrained generative architecture to generate new \emph{well-formed} causal/non-causal sentences that contain them.
	

	\begin{table*}[h] \footnotesize
		\centering
		\scalebox{0.90}{
		\begin{tabular}{|m{1.5cm}|m{9cm}|m{5.5cm}|}
			\hline
			\multicolumn{1}{|c|}{\textbf{Knowledge}}  & \multicolumn{1}{c|}{\textbf{How to extract event pair}}    & \multicolumn{1}{c|}{\textbf{Why causal or non-causal}}   \\ \hline
			\multicolumn{3}{|c|}{\textbf{Lexical knowledge expanding}}      \\ \hline
			\multicolumn{1}{|c|}{\textbf{WordNet}}                                                                    & 1) Extracting the synonyms and hypernyms from WordNet of each event in $ep$. 2) Assembling the items from the two groups of two events to generate causal/non-causal event pairs.    & Items in each group are the synonyms and hypernyms of the annotated causal/non-causal event pairs.  \\ \hline
			\multicolumn{1}{|c|}{\textbf{VerbNet}}                                                                    & 1) Extracting the words from VerbNet under the same class as each event in $ep$. 2) Assembling the items from the two groups of two events to generate causal/non-causal event pairs.     & Items in each group are in the same class of the annotated causal/non-causal event pairs.         \\ \hline
			
			\multicolumn{1}{|c|}{\textbf{e.g.}}  & \multicolumn{2}{c|}{$<(killed,attack),causal>\Longrightarrow kill \stackrel{Synonyms}{\longrightarrow}hurt$, $attack \stackrel{Synonyms}{\longrightarrow}onrush\Longrightarrow <(hurt,onrush),causal>$} \\ 
			\multicolumn{1}{|c|}{}  & \multicolumn{2}{c|}{\emph{Original sentence}: Kimani Gray, a young man who likes football, was killed in a police attack shortly after a tight match.} \\ \hline
			
			\multicolumn{3}{|c|}{\textbf{Connective knowledge introducing}}    \\ \hline
			\textbf{FrameNet PDTB2} & 1) Extracting causal/non-causal connectives from FrameNet\footnote{Frame with types of \emph{Reasoning}, \emph{Causation}, \emph{Causation\_scenario}, \emph{Reason}, \emph{Triggering} and \emph{Explaining\_the\_facts}.} and PDTB2. 2) Extracting any two events connected by causal/non-causal connectives on KBP corpus to obtain causal/non-causal event pairs and original sentences respectively.     & Introduced event pairs are connected by causal/non-causal connectives.      \\ \hline
			\multicolumn{1}{|c|}{\textbf{e.g.}}  & \multicolumn{2}{c|}{\textbf{Looting} \emph{because} someone \textbf{beat up} someone, like the Travon Martin case. $\stackrel{because}{\Longrightarrow}<(loot,beat\_up),causal>$}  \\ 
			\multicolumn{1}{|c|}{}  & \multicolumn{2}{c|}{\emph{Original sentence}: Looting because someone beat up someone, like the Travon Martin case.} \\ \hline
			
		\end{tabular}}
		\caption{Extracting causal and non-causal event pairs from multiple knowledge bases.}
		\label{tab1}
	\end{table*}
	
	\paragraph{Knowledge Guiding} 
	\label{sec:KG} 
		\label{sec:KSG}
	
	KSG introduces event pairs that are probabilistic causal or non-causal from multiple knowledge bases in two ways. (1) \emph{Lexical knowledge expanding}: expanding annotated event pairs via external dictionaries, such as WordNet \cite{miller1995wordnet} and VerbNet \cite{schuler2005verbnet}. (2) \emph{Connective knowledge introducing}: introducing event pairs from external event-annotated documents (KBP corpus) assisted with FrameNet \cite{Baker1998TheBF} and Penn Discourse Treebank (PDTB2) \cite{pdtb2008pdtb}. As shown in Table \ref{tab1}, we illustrate how to extract event pairs from multiple knowledge bases. Then, inspired by \citet{bordes2013translating}, we filter the extracted event pairs by converting them into triples $<$$e_i$, causal/non-causal, $e_j$$>$ and calculating the causal-distance by maximizing $L$ in a causal representation space:
	\begin{equation}\footnotesize
	L=\sum_{(e_i,e_j) \in T} \sum_{(e'_i,e'_j) \in T'} [\lambda + d(\bm{e'_i},\bm{e'_j}) - d(\bm{e_i},\bm{e_j})]_{+}, 
	\end{equation}
	where $T$ and $T'$ are the causal and non-causal triples set respectively, and $\bm{e}$ is the representation of event. After that, the higher probability of causal relation, the shorter distance between two events, and we sort event pairs in ascending order by their distances. Finally, we keep the top and bottom $\alpha$\% sorted event pairs to obtain the causal and non-causal event pairs sets for generation.
	
	\paragraph{Constrained Sentence Generator} Given an event pair, constrained sentence generator produces a well-formed sentence that expresses its causal or non-causal relation in three stages: (1) \emph{assigning event-related entities} ensures the logic of the semantic roles of events, (2) \emph{completing sentences} ensures the completeness of causal or non-causal semantic expression, (3) \emph{filtering sentences} ensures the quality and diversity of generated sentences.
	
	\emph{\textbf{Assigning Event-related Entities.}} Event related entities play different semantic roles of events in sentences, which is an important part of event-semantic expression. Hence, as shown in Figure \ref{fig4}, given an event pair, we firstly assign logical entities for input events to guarantee the logic of semantic roles in the new sentences, such as \emph{gang} is a logical entity as the body of the event \emph{onrush}.
	Logically, entities of the same type play the same semantic roles in similar events. Moreover, as shown in Table \ref{tab1}, there is a corresponding original sentence for each extracted event pair. 
	Therefore, in new sentence, we assign the most similar entity in the same type from candidate set\footnote{We collect entities from annotated data and KBP corpus.} for each entity in the original sentence. For example, we assign \emph{gang} for \emph{onrush} in new sentence which is similar with the \emph{police} related to \emph{attack} in the original sentence. Specifically, we put the candidate entities in the same position in the original sentence to obtain their BERT embeddings. Then we select entities via the cosine similarity between their embeddings: $\mathcal{E}(ent) = \frac{1}{|ent|}\sum_{w \in ent}\mathcal{E}(w)$, where $ent$ is the entity and $\mathcal{E}(w)$ is the BERT embedding of $ent$.
	
	\emph{\textbf{Completing Sentences.}} A well-formed sentence requires a complete linguistic expression to express the causal or non-causal semantics. Therefore, we complete sentences by filling the cohesive words between given events and assigned entities with masked BERT \cite{devlin-etal-2019-bert}. All words except events and entities are regarded as cohesive words. Specifically, we insert a certain number of the special token [MASK] between events and entities, and then predict the [MASK]\footnote{The inserted [MASK] is 1.2 times the number of words between events and entities in the original sentence.} tokens as new words. As shown in Figure \ref{fig4}, we fill cohesive tokens via two independent generators to express causal and non-causal semantic according to the relation of given events. For example, \emph{in a} guiding a causal semantic filled by the causal generator. 
	
	\emph{\textbf{Filtering Sentences.}} Inspired by \citet{yang-etal-2019-exploring-pre}, we design a filter to select new sentences that are balanced between high quality and high diversity with two key factors: 1) \textbf{Perplexity} (PPL): we take the average probability of the filled cohesive words in the new sentence $s'$ as its perplexity: $PPL(s') = \frac{1}{|T(s')|} \sum_{t \in T(s')} P(t)$, where $T$ is the set of filled cohesive words.
	2) \textbf{Distance} (DIS): we calculate the cosine similarity between generated sentence $s'$ and annotated data $D_m$ as its distance:
	$DIS(s',D_m)=\frac{1}{|D_m|} \sum_{s \in D_m} \frac{\mathcal{E}(s') \cdot \mathcal{E}(s)}{\mathcal{E}(s') \times \mathcal{E}(s)}$, where $D_m$ is $m$ random selected annotated sentences and $\mathcal{E}$ is the BERT sentence representation of the [CLS] token. A new sentence should have both appropriate high PPL which indicates the quality of generation, and appropriate high DIS which indicates the difference from the original sentences. Therefore, we select the top $\beta$\% of the newly generated sentences according to $Score$ for the further training of identifier as following: $Score(s')=\mu PPL(s') + (1- \mu) DIS(s',D_m))$, where the $\mu$ is an hyper-parameter. 
	
	\subsection{Training of LearnDA for ECI}
	\label{sec:Training}
	We briefly describe the training processes of LearnDA for ECI, including the pre-training of generator and identifier, the dual reinforcement training, and the further training of identifier.
	
	\begin{algorithm}[t] \footnotesize
		\caption{Dual Reinforcement Training of $\mathcal{G}$ $\mathcal{I}$.}
		\begin{algorithmic}[1]
			\Require A set of knowledge guided event pairs \{($ep$,$s$,$c$)\}

			A pre-trained generator $\mathcal{G}$ and identifier $\mathcal{I}$ 
			\Ensure Early stop on the development set according to $\mathcal{I}$.
			
			\Function {Primal Cycle}{}
			\For{event pair $(ep_i,s_i,c_i)$ in batch}
			\State Generator generates the sentence $s'_i$ of $ep_i$;
			\State Identifier re-predicts the causality $c^*_i$ of $ep_i$;
			\State Computing the reward as:
			\State $\quad R_{primal}^s = \lambda R_s(ep_i,c_i)+(1-\lambda) R_c(ep_i,s'_i)$.
			\State Computing the stochastic gradient of $\theta_{\mathcal{G}}$:
			\State $\quad \nabla_{\mathcal{G}} += R_{primal}^s \cdot \nabla_{\theta_{\mathcal{G}}} L_{G}(ep_i, c_i)$.
			\EndFor
			
			\State Model batch updates: $\theta_{\mathcal{G}} \leftarrow \theta_{\mathcal{G}} + \eta \cdot \nabla_{\mathcal{G}}$
			\EndFunction  
			\\
			\Function {Dual Cycle}{}
			\For{event pair $(ep_i,s_i,c_i)$ in batch}
			\State Identifier predicts the causality $c'_i$ of $ep_i$;
			\State Generator re-generates the sentence $s^*_i$ of $ep_i$;
			\State Computing the reward as:
			\State $\quad R_{dual}^s = \gamma R_c(ep_i,s_i)+(1-\gamma) R_s(ep_i,c'_i)$.
			\State Computing the stochastic gradient of $\theta_{\mathcal{I}}$:
			\State $\quad \nabla_{\mathcal{I}} += R_{dual}^s \cdot \nabla_{\theta_{\mathcal{I}}} L_{I}(ep_i, s_i)$.
			\EndFor
			
			\State Model batch updates: $\theta_{\mathcal{I}} \leftarrow \theta_{\mathcal{I}} + \eta \cdot \nabla_{\mathcal{I}}$
			\EndFunction
			
		\end{algorithmic}
		\label{alg1}
	\end{algorithm}
	
	\paragraph{Event Causality Identifier}
	\label{sec:detector} First of all, we formulate event causality identification as a sentence-level binary classification problem. Specifically, we design a classifier based on BERT \cite{devlin-etal-2019-bert} to build our identifier. The input of the identifier is the event pair $ep$ and its sentence $s$. Next, we take the stitching of manually designed features (same lexical, causal potential, and syntactic features as \citet{gao-etal-2019-modeling}) and two event representations as the input of top MLP classifier. Finally, the output is a binary vector to predict the causal/non-causal relation of the input event pair $ep$.
	
	\paragraph{Pre-training} We pre-train the identifier and generator on labeled data before dual reinforcement training. On the one hand, we train identifier via the cross-entropy objective function of the relation classification. On the other hand, for generators, we keep the events and entities in the input sentences, replace the remaining tokens with a special token [MASK], and then train it via the cross-entropy objective function to re-predict the masked tokens. Specifically, causal generator and non-causal generator are pre-trained on causal and non-causal labeled sentences respectively.
	
	\paragraph{Dual Reinforcement Training} As shown in Algorithm \ref{alg1}, we interactively optimize the generator and identifier by dual reinforcement learning. Specifically, we maximize the following objective functions:
	\begin{equation}\footnotesize
	\setlength{\abovedisplayskip}{6pt} 
	\setlength{\belowdisplayskip}{6pt}
	L_{G}(ep, c) = 
	\begin{cases}
	p(s'|c;\theta_{G}) = \frac{1}{|T_s|} \sum_{t \in T_s} p(t|c;\theta_{G}) \\
	p(s'|c;\theta_{NG}) = \frac{1}{|T_s|} \sum_{t \in T_s} p(t|c;\theta_{NG}),
	\end{cases} 
	\end{equation}
	\begin{equation}\footnotesize 
	\setlength{\abovedisplayskip}{3pt} 
	\setlength{\belowdisplayskip}{3pt}
	L_{I}(ep, s) = p(c'|s;\theta_{I}),
	\end{equation}
	where $\theta_{G}$ and $\theta_{NG}$ is the parameters of causal and non-causal sentence generators respectively, $T_s$ is the masked tokens. Finally, after dual data augmentation, we utilize generated sentences to further train the dual-trained identifier via the cross-entropy objective function of relation classification. 
	
	\section{Experiments}
	\subsection{Experimental Setup}
	\paragraph{Dataset and Evaluation Metrics} 
	Our experiments are conducted on two main benchmark datasets, including: \textbf{EventStoryLine} v0.9 (ESC) \cite{caselli2017event} described above; and (2) \textbf{Causal-TimeBank} (Causal-TB) \cite{Mirza2014AnAO} which contains 184 documents, 6813 events, and 318  causal event pairs. Same as previous methods, we use the last two topics of ESC as the development set for two datasets. For evaluation, we adopt Precision (P), Recall (R), and F1-score (F1) as evaluation metrics. We conduct 5-fold and 10-fold cross-validation on ESC and Causal-TB respectively, same as previous methods to ensure comparability.
	All the results are the average of three independent experiments. 
	
	\paragraph{Parameters Settings} In implementations, both the identifier and generators are implemented on BERT-Base architecture\footnote{\url{https://github.com/google-research/bert}}, which has 12-layers, 768-hiddens, and 12-heads. We set the learning rate of generator pre-training, identifier pre-training/further training, and dual reinforcement training as 1e-5, 1e-5, and 1e-7 respectively. We set the ratio of the augmented data used for training to the labeled data, $\alpha$, $\beta$, $\mu$, $\lambda$ and $\gamma$ as 1:2, 30\%, 50\%, 0.2, 0.5 and 0.5 respectively tuned on the development set. And we apply early stop and SGD gradient strategy to optimize all models. We also adopt a negative sampling rate of 0.5 for training the identifier, owing to the sparseness of positive examples. (See Appendix D for more details.)
	
	\paragraph{Compared Methods} Same as previous state-of-the-art work. For ESC, we prefer 1) \textbf{LSTM } \cite{cheng2017classifying}, a dependency path based sequential model that models the context between events to identify causality; 2) \textbf{Seq} \cite{choubey2017sequential}, a sequence model explores complex human designed features for ECI; 3) \textbf{LR+} and \textbf{ILP} \cite{gao-etal-2019-modeling}, document-level models adopt document structures for ECI. For Causal-TB, we prefer 1) \textbf{RB}, a rule-based system; 2) \textbf{DD}, a data driven machine learning based system; 3) \textbf{VR-C}, a verb rule based model with data filtering and gold causal signals enhancement. These models are designed by \citet{Mirza2014AnAO,DBLP:journals/corr/Mirza16} for ECI. 
	
	Owing to our methods are constructed on BERT, we build BERT-based methods: 1) \textbf{BERT}, a BERT-based baseline, our basic proposed event causality identifier. 2) \textbf{MM} \cite{ijcai2020-499}, the BERT-based SOTA method with mention masking generalization. 3) \textbf{MM+$Aug$}, the further re-trained MM with our dual augmented data. 4) \textbf{KnowDis} \cite{zuo-etal-2020-knowdis} improved the performance of ECI with the distantly labeled training data. We compare with it to illustrate the quality of our generated ECI-related training data. 5) \textbf{MM}+$ConceptAug$, to make a fair comparison, we introduce causal-related events from ConceptNet that employed by MM, and generate new sentences via KonwDis and LearnDA to further re-train MM (see Appendix C for details). Finally, we use \textbf{LearnDA}$_{Full}$ indicates our full model, which is the dual-trained identifier further trained via dual augmented data.
	
	
	\subsection{Our Method vs. State-of-the-art Methods}
	\begin{table}[t] \footnotesize
		\centering
		\begin{tabular}{lccc}
			\textbf{Methods}    & \textbf{P} & \textbf{R} & \textbf{F1}  \\ \hline
			\multicolumn{4}{c}{\textbf{ESC}} \\ \hline
			LSTM \cite{cheng2017classifying}         & 34.0       & 41.5       & 37.4  \\ 
			Seq \cite{choubey2017sequential}         & 32.7       & 44.9       & 37.8 \\ 
			LR+ \cite{gao-etal-2019-modeling}        & 37.0       & 45.2       & 40.7  \\ 
			ILP \cite{gao-etal-2019-modeling}        & 37.4       & 55.8       & 44.7   \\ 
			BERT      & 36.1       & 56.0       & 43.9       \\ 
			KnowDis \cite{zuo-etal-2020-knowdis} &  39.7 &  66.5  &  49.7  \\
			MM \cite{ijcai2020-499} &  41.9 &  62.5  &  50.1  \\ \hline
			MM+${ConceptAug}$ (\textbf{Ours}) &  41.2 &  66.5  &  50.9* \\
			MM+${Aug}$ (\textbf{Ours}) &  41.0 &  69.3  &  51.5* \\
			\textbf{LearnDA}$_{Full}$ (\textbf{Ours})     & \textbf{42.2}         & \textbf{69.8}        & \textbf{52.6*}   \\ \hline
			\multicolumn{4}{c}{\textbf{Causal-TB}}    \\ \hline
			RB \cite{Mirza2014AnAO}  &  36.8  &  12.3  & 18.4  \\ 
			DD \cite{Mirza2014AnAO}  &  67.3  &  22.6  & 33.9  \\
			VR-C \cite{DBLP:journals/corr/Mirza16}  &  \textbf{69.0}  &  31.5  & 43.2    \\ 
			BERT     & 38.5  & 43.9  &  41.0   \\ 
			MM \cite{ijcai2020-499}    & 36.6  &  55.6  &  44.1  \\
			KnowDis \cite{zuo-etal-2020-knowdis}    & 42.3  &  60.5  &  49.8  \\ \hline
			MM+${ConceptAug}$ (\textbf{Ours})   & 38.8  & 59.2  &  46.9*  \\ 
			MM+${Aug}$ (\textbf{Ours})   & 39.2  & 61.9  &  48.0*  \\ 
			\textbf{LearnDA}$_{Full}$ (\textbf{Ours})   &   41.9    &  \textbf{68.0}   &  \textbf{51.9*}  \\ \hline
		\end{tabular}
		\caption{Results on event causality identification. * denotes a significant test at the level of 0.05.} 
		\label{tab2} 
	\end{table}
	
	Table \ref{tab2} shows the results of ECI on EventStoryLine and Causal-TimeBank. From the results:
	
	1) Our LearnDA$_{Full}$ outperforms all baselines and achieves the best performance (52.6\%/51.9\% on F1 value), outperforming the no-bert (ILP/VR-C) and bert (MM/KnowDis) state-of-the-art methods by a margin of 7.9\%/8.7\% and 2.5\%/2.1\% respectively, which justifies its effectiveness. Moreover, BERT-based methods demonstrate high recall value, which is benefited from more training data and their event-related guided knowledge.
	
	2) Comparing KnowDis with LearnDA$_{Full}$, we note that training data generated by LearnDA is more helpful to ECI than distant supervision with external knowledge (+2.9\%/+2.1\%). This shows that LearnDA can generate more ECI-related data.
	
	3) Comparing MM+$ConceptNet$ with MM, with the same knowledge base, our dual augmented data can further improve the performance (+0.8\%/+2.8\%), which illustrates that LearnDA can make more effective use of external knowledge by generating task-related training data.
	
	4) Comparing MM+${Aug}$ with MM, we note that training with our dual augmented data can improve the performance by 1.4\%/3.9\%, even though MM is designed on BERT-Large (LearnDA is constructed on BERT-Base) and also introduces external knowledge. This indicates that the augmented data generated by our LearnDA can effectively alleviate the problem of data lacking on the ECI.

	\begin{table}[t] \footnotesize
		\centering
		\begin{tabular}{cccc}
			\multicolumn{1}{l}{\textbf{Method}}         & \multicolumn{1}{c}{\textbf{P}} & \multicolumn{1}{c}{\textbf{R}} & \multicolumn{1}{c}{\textbf{F}}      \\ \hline
			\multicolumn{1}{l}{BERT (Our basic identifier)}    & 36.1       & 56.0       & 43.9                              \\
			\multicolumn{1}{l}{BERT$_{OrgAug}$}         & 36.6       & 59.7       & 45.4*                         \\
			\multicolumn{1}{l}{BERT$_{DualAug}$}                     & 37.8                            & 65.6                            & 48.0*                         \\
			\multicolumn{1}{l}{LearnDA$_{Dual}$}                          & 36.8                            & 63.0                            & 46.5*                                 \\
			\multicolumn{1}{l}{LearnDA$_{DualAug-w/o.KB}$}                            & 37.5                            & 67.0                            & 48.1*                              \\
			\multicolumn{1}{l}{$-$LearnDA$_{DualAug-w/.intro}$}                             & 39.0                            & 66.0                            & 49.0*                                    \\ 
			\multicolumn{1}{l}{$-$LearnDA$_{DualAug-w/.verbnet}$}         &                39.4                            & 66.7                            & 49.5*                            \\ 
			\multicolumn{1}{l}{$-$LearnDA$_{DualAug-w/.wordnet}$}            & 39.6         & 67.6        & 49.9*                      \\ 
			\multicolumn{1}{l}{LearnDA$_{Full}$} & 42.2         & 69.8        & \textbf{52.6*}  \\ \hline
		\end{tabular}
		\caption{Ablation results on event causality identification on ESC. * denotes a significant test at the level of 0.05. BERT$_{OrgAug}$ and BERT$_{DualAug}$ denote the BERT is further trained on no-dual and dual augmented data respectively; LearnDA$_{Dual}$ denotes our identifier is only trained by dual learning without further training; LearnDA$_{DualAug-w/o.KB}$ denotes the LearnDA$_{Dual}$ is further trained by dual augmented data without knowledge guiding; LearnDA$_{DualAug-w/.<kb>}$ denotes LearnDA$_{Dual}$ is further trained by dual augmented data guided with knowledge base $kb$.}
		\label{tab3}
	\end{table}
	
	\subsection{Effect of Learnable Dual Augmentation}
	We analyze the effect of the learnable dual augmentation for event causality identification. 1) \emph{For identifier}. Comparing LearnDA$_{Dual}$ with BERT in Table \ref{tab3}, we note that the performance of the proposed identifier is improved (+2.6\%) after the dual training only with the same labeled data. This indicates that the identifier can learn more informative expressions of causal semantic from generation with dual learning. 2) \emph{For generator}. Comparing BERT$_{DualAug}$ with BERT$_{Aug}$ in Table \ref{tab3}, we note that the dual augmented data is high quality and more helpful to ECI (+2.6\%). This indicates generator can generate more ECI task-related data learned from identifier with dual learning. 
	
	\begin{figure}[t] 
		\centering
		\includegraphics*[clip=true,width=0.35\textwidth,height=0.18\textheight]{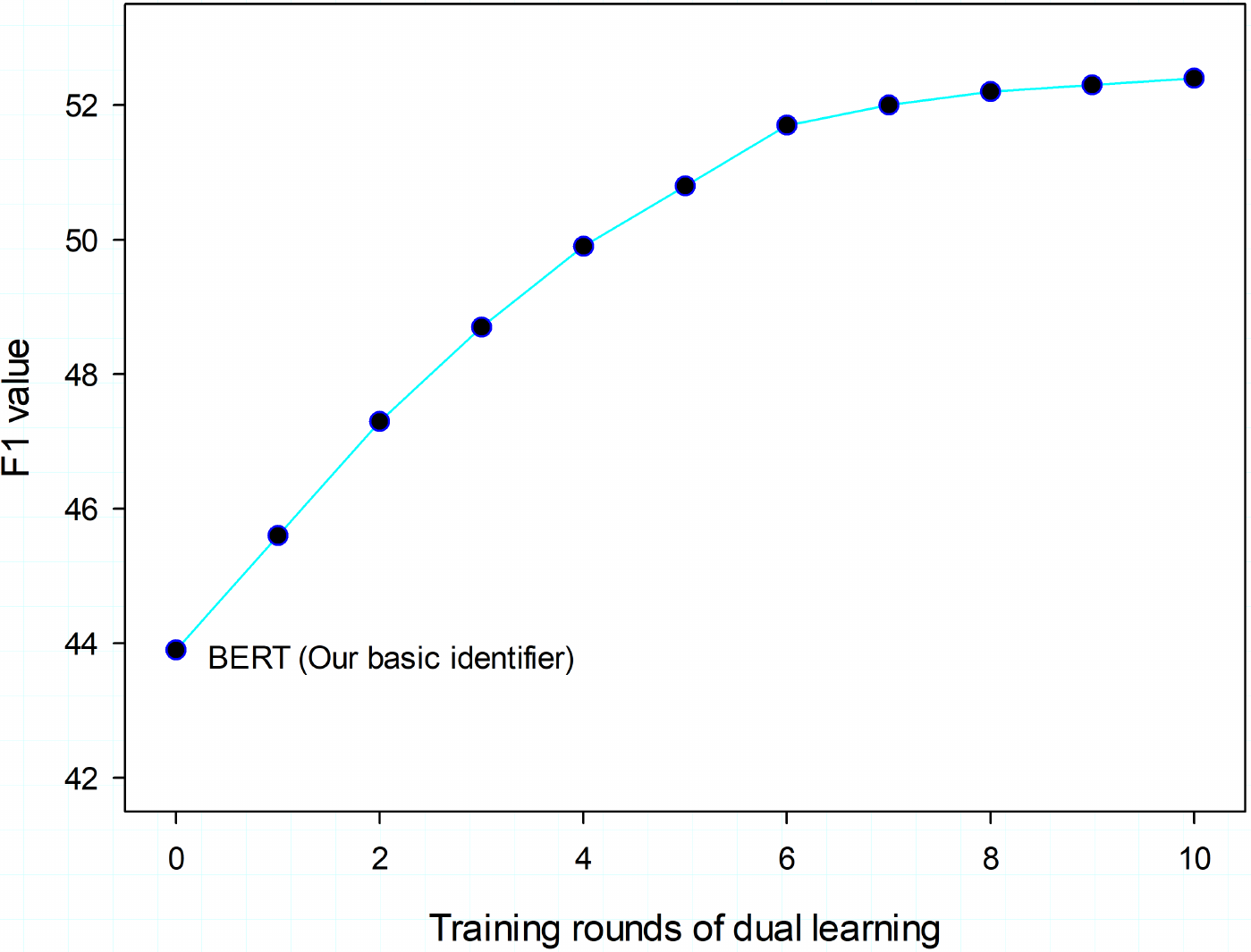}
		\caption{The impact of the training rounds of dual learning on event causality identification on ESC. In each round, we generate new training data by the generator at the current round. The performance is achieved by further training the identifier at the current round with the aforementioned newly generated data.} \label{fig5}
	\end{figure}
	
	Figure \ref{fig5} illustrates the learnability of our LearnDA. Specifically, as the number of training rounds of dual learning increases, the generated data gradually \emph{learns} task-related information, further improving the performance accordingly.
	
	\subsection{Effect of Knowledge Guiding}
	Table \ref{tab3} also illustrates the effect of knowledge guiding on ECI depending on different knowledge bases. 1) Comparing LearnDA$_{Full}$ with LearnDA$_{DualAug-w/o.KB}$, we note that the augmented data guided by external knowledge can further improve the performance of ECI. 2) Specifically, \emph{lexical expanding} and \emph{connective introducing} (Sec \ref{sec:KG}) can both make the representation of causal relation more generalized, further making it easier for the identifier to understand the causality. 3) Moreover, the expanding is more effective than the introducing, because the former brings a wider range of effective knowledge, thus the guidance of causal-related knowledge is better.

	\subsection{Our Augmentation vs. Other NLP Augmentations}
		\begin{table}[t] \footnotesize
		\centering
		\begin{tabular}{cccc}
			\multicolumn{1}{l}{\textbf{Method}}         & \multicolumn{1}{c}{\textbf{P}} & \multicolumn{1}{c}{\textbf{R}} & \multicolumn{1}{c}{\textbf{F}}      \\ \hline
			\multicolumn{1}{l}{BERT (Our identifier)}    & 36.1       & 56.0       & 43.9                              \\
			\multicolumn{1}{l}{TextSurface$_{BERT}$}                            & 37.0                            & 57.5                            & 45.0*                              \\
			\multicolumn{1}{l}{BackTranslation$_{BERT}$}                          & 36.8                            & 61.0                            & 45.9*                                 \\
			\multicolumn{1}{l}{EDA$_{BERT}$}                     & 36.6                            & 62.4                            & 46.1*                         \\
			\multicolumn{1}{l}{LearnDA$_{BERT}$}                     & 37.8                            & 65.6                            & \textbf{48.0*}           \\ \hline
		\end{tabular}
		\caption{Results of different data augmentation methods on event causality identification on ESC dataset. * denotes a significant test at the level of 0.05.}
		\label{tab4}
	\end{table}
	In this section, we conduct a comparison between our augmentation framework and other NLP-related augmentation methods to further illustrate the effectiveness of LearnDA.
	
	\paragraph{Effectiveness of Our Augmentation}
	We train our identifier with augmented data produced by different NLP-related augmentation methods. As shown in Table \ref{tab4}, the augmented data generated by our LearnDA is more efficient for ECI, which is consistent with the previous analysis. The LearnDA can generate well-formed task-related new sentences that contain more event causal knowledge. Specifically, 1) \emph{text surface transformation} brings a slight change to the labeled data, thus it has relatively little impact on ECI; 2) \emph{Back translation} introduces limited new causal expressions by translation, thus it slightly increases the recall value on ECI; 3) \emph{EDA} can introduce new expressions via substitution, but the augmented data is not canonical and cannot accurately express the causality, therefore, its impact on ECI is also limited. 
	
		\begin{table}[t] \footnotesize 
		\centering
		\resizebox{0.47\textwidth}{!}{
			\begin{tabular}{lcccc}
				\textbf{}                & \multicolumn{1}{c}{\textbf{Gold}} & \multicolumn{1}{c}{\textbf{EDA}} & \multicolumn{1}{c}{\textbf{BackTrans}} & \multicolumn{1}{c}{\textbf{LearnDA}} \\ \hline
				\textbf{Causality}  &   3.80   & 3.20 &   3.70  & 3.60   \\
				\textbf{Well-formedness} &  3.95  &  2.75  &  3.83  & 3.64  \\
				\textbf{Diversity} (Man/Auto) & 0.0/1.0  &  3.08/0.70  & 2.80/0.85  & 3.51/0.66 \\ \hline
		\end{tabular}}
		\caption{Manual (4-score rating (0, 1, 2, 3)) and automatic (BLEU score) evaluation of the generated sentences via different methods from causality, well-formedness and diversity. Causality and well-formedness are assessed manually, while diversity is assessed manually and automatically.} 
		\label{tab5} 
	\end{table}
	
	\paragraph{Quantitative Evaluation of Task-relevance}
	We select five Ph.D. students majoring in NLP to manual score the 100 randomly selected augmented sentences given their corresponding original sentences as reference (Cohen's kappa = 0.85). Furthermore, we calculate the BLEU \cite{Papineni2002BleuAM} value to further evaluate the diversity. As aforementioned, the task-relevance of new sentences on ECI is manifested in causality and well-formedness, while the diversity indicates the degree of generalization. As shown in Table \ref{tab5}, we note the sentences generated by LearnDA are equipped with the above three properties that are close to the labeled sentences. Specifically, the sentences produced by EDA has a certain degree of causality and diversity due to the lexical substitution assisted by external knowledge. However, they cannot well express the causality due to the grammatical irregularities. Correspondingly, new sentences generated via back translation are very similar to the original sentences, while the diversity is poor.
	
	\subsection{Case Study}
	\begin{figure}[t] 
		\centering
		\includegraphics*[clip=true,width=0.44\textwidth,height=0.18\textheight]{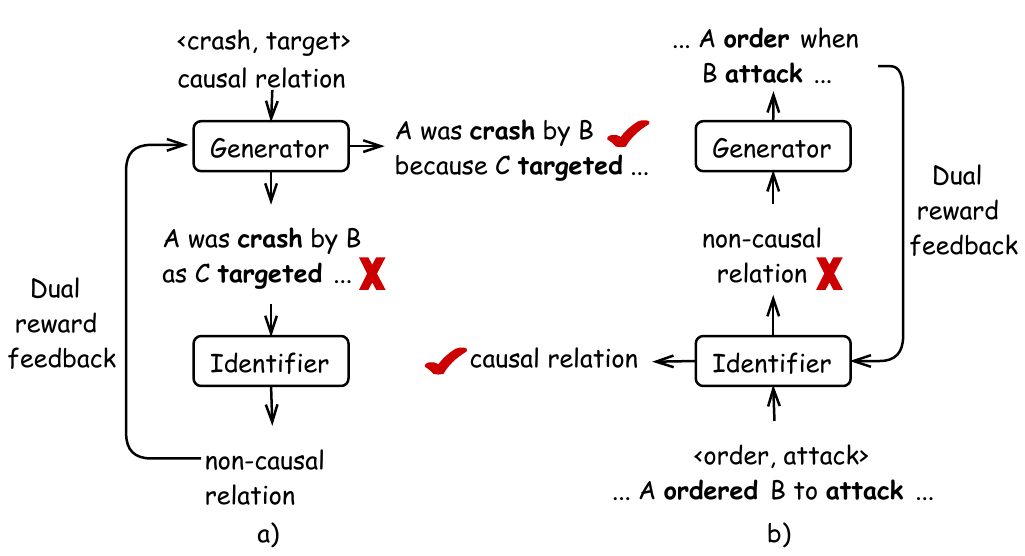}
		\caption{The modification of dual learning.} \label{fig6}
	\end{figure}
	We conduct a case study to further investigate the effectiveness of our LearnDA. Figure \ref{fig6} illustrates the modification process of dual learning. For example as a), given two causal events, the generator is expected to generate a causal sentence. However, the generator without dual learning produces a non-causal sentence. Fortunately, with dual learning, the identifier judges the generated sentence as a non-causal one and guides the generator to produce a causal sentence with the feedback. Similarly, as shown in b), given a causal sentence, the identifier is expected to output a causal relation, but no dual-trained one cannot do. Correspondingly, the generator constructs feedback of low confidence to guide the identifier to output a causal relation.
	
	\section{Conclusion}
	This paper proposes a new learnable knowledge-guided data augmentation framework (LearnDA) to solve the data lacking problem on ECI. Our framework can leverage the duality between generation and identification via dual learning to generate task-related sentences for ECI. Moreover, our framework is knowledge guided and learnable. 
	Our method achieves state-of-the-art performance on EventStoryLine and Causal-TimeBank datasets. 

\section*{Acknowledgments}
We thank anonymous reviewers for their insightful comments and suggestions. This work is supported by the National Key Research and Development Program of China (No.2018YFB1005100), the National Natural Science Foundation of China (No.U1936207, 61806201). This work is also supported by Beijing Academy of Artificial Intelligence (BAAI2019QN0301) and the joint project with Beijing Baidu Netcom Science Technology Co., Ltd.

\bibliographystyle{acl_natbib}
\bibliography{anthology,acl2021}

\newpage
\appendix
	\section{Supplementary Experiment Results}
	
	\subsection{Statistics of Dual Augmented Data}
	\begin{table}[h] \footnotesize
		\centering
		\begin{tabular}{lcc}
			\hline
			& \textbf{Annotated data} & \textbf{Augmented data}  \\ \hline                                                                      
			\#causal ep. & 1170 & 3588 \\
			\#causal sent. & 1770 & 10442 \\
			\#Ave sent. &  1.5 & 2.9 \\ \hline
		\end{tabular}
		\caption{Statistics of causal event pairs and causal sentences in labeled data (ESC) and dual augmented data. (\textbf{\#causal ep.} denotes the number of causal event pairs after removing duplicates, \textbf{\#causal sent.} denotes the number of causal sentences, \textbf{\#Ave sent.} denotes the average number of causal sentences containing the same causal event pair.)} 
		\label{tab6} 
	\end{table}
	
	As shown in Table \ref{tab6}, our dual augmented data is significantly more quantitative than the labeled data. Specifically, the causal event pairs are increased by 3.1 times, the causal sentences are increased by 5.9 times and the average number of causal sentences corresponding to each causal event pair is also increased.
	
	\subsection{Effectiveness of Different Quantities of Augmented Training Data} 
		\begin{table}[h] \footnotesize
		\centering
		\begin{tabular}{ccccc}
			\hline
			\textbf{Ratio}      & \textbf{P} & \textbf{R} & \textbf{F1}    \\ \hline
			1:1 & 37.3       & 64.7       & 47.3*     \\ 
			1:2 & 37.8       & 65.6      & \textbf{48.0}*  \\ 
			1:3 & 37.0       & 64.8      & 47.1*          \\ 
			1:4 & 36.2       & 64.2       & 46.3*           \\ \hline
		\end{tabular}
		\caption{Performance of identifier (BERT) trained with different ratios of labeled data and dual augmented data. * denotes a significant test at the level of 0.05.}
		\label{tab7}
	\end{table}
	We change the quantity of dual augmented data for training to explore the influence of augmentation ratio on ECI. As shown in Table \ref{tab7}, when the ratio is 1:2, the effective knowledge brought by dual augmented data is maximized. And as the ratio increasing, the dual augmented data will bring noises, which obstructs the model to identify event causality and may change the data distribution from original data \cite{Xie2019AdversarialEI}. This suggests that too much augmented data is not better and that there is a trade-off between introducing knowledge and reducing noise.
	
	\subsection{Effectiveness of Extracting Event Pairs with Different Filtering Ratios} 
	\begin{table}[h] \footnotesize
		\centering
		\begin{tabular}{ccccc}
			\hline
			$\bm{\alpha}$ & \textbf{P} & \textbf{R} & \textbf{F1}   & \multicolumn{1}{l}{\textbf{$\bm{\nabla}$}} \\ \hline
			30\% & 37.8       & 65.6      & \textbf{48.0}* & \textbf{-}                                  \\ 
			40\% & 37.0       & 65.7       & 47.3*          & -0.7                                        \\ 
			50\% & 36.2       & 65.0       & 46.5*          & -1.5                                        \\ \hline
		\end{tabular}
		\caption{Performance of identifier (BERT) trained with different extracting event pairs filtered in different $\alpha$. * denotes a significant test at the level	of 0.05.}
		\label{tab8}
	\end{table}
	Table \ref{tab8} tries to show the effectiveness of extracting event pairs with different filtering ratios on ECI. With the ratio of retained event pairs increasing, the augmented data hurts ECI's performance. This proves the effectiveness of filtering, which further improves the causality of the generated sentences.
	
	\subsection{Effectiveness of Generated sentences with Different Filtering Ratios} 
	\begin{table}[h] \footnotesize
		\centering
		\begin{tabular}{cccccc}
			\hline
			$\bm{\beta}$           & \textbf{P} & \textbf{R} & \textbf{F1}  & \multicolumn{1}{l}{\textbf{$\bm{\nabla}$}}   \\ \hline
			50\% & 37.8       & 65.6      & \textbf{48.0}* & - \\ 
			60\% & 37.3       & 65.3       & 47.5*     &  -0.5   \\ 
			70\% & 36.9       & 64.9       & 47.0*     &  -1.0   \\ 
			80\% & 36.6       & 64.5       & 46.7*     &  -1.3  \\ \hline
		\end{tabular}
		\caption{Performance of identifier (BERT) trained with new generated sentences filtered in different $\beta$. * denotes a significant test at the level of 0.05.}
		\label{tab9}
	\end{table}
	Table \ref{tab9} tries to show the effectiveness of generated sentences with different filtering ratios. With the ratio of retained generated sentences increasing, the contribution of filtered generated sentences for ECI decreases gradually. This proves the effectiveness of filtering, which can balance the overall quality of the sentences against diversity.
	
	\section{Supplementary Related Work}
	
	\subsection{Dual Learning}
	
	For many Natural Language Processing (NLP) tasks, there exist many primal and dual tasks, such as open information narration (OIN) and open information extraction (OIE) \cite{Sun2018LogicianAO}, natural language understanding (NLU) and natural language generation (NLG) \cite{su-etal-2019-dual,su-etal-2020-towards}, semantic parsing and natural language generation \cite{Ye2019JointlyLS,Cao2019SemanticPW,cao-etal-2020-unsupervised}, link prediction and entailment graph induction \cite{Cao2019SemanticPW}, query-to-response and response-to-query generation \cite{shen-feng-2020-cdl} and so on. The duality between the primal task and the dual task is considered as a constraint that both problems must share the same joint probability mutually. Recently, inspired by \citet{xia2017dual} who implemented the duality in a neural-based dual learning system, the above primal-dual tasks are implemented in two different ways: 1) providing additional labeled samples via bootstrapping, and 2) adding rewards at the training stage for each agent. We observe that the event causality identification and the sentence generation are dual to each other. Therefore, we apply a dual learning framework in the second way to optimize identification and generation interactively for generating ECI-related data. 
	
	\subsection{Data Augmentation for NLP} 
	The scarcity of annotated data is a thorny problem in machine learning. Unlike computer vision, the augmentation of text data in NLP is pretty rare. Existing text data augmentation methods for NLP tasks are almost task-independent frameworks and can be roughly summarized into the following categories \cite{chaudhary2020nlpaugment}: (1) Lexical substitution tries to substitute words without changing the meaning \cite{Zhang2015CharacterlevelCN,wei-zou-2019-eda,wang-yang-2015-thats,Xie2019UnsupervisedDA}; (2) Back translation tries to paraphrase a text while retraining the meaning \cite{Xie2019UnsupervisedDA}; (3) Text surface transformation tries to match transformations using regex \cite{Coulombe2018TextDA}; (4) Random noise injection tries to inject noise in the text to make the model more robust \cite{wei-zou-2019-eda}; (5) Generative method tries to generate additional training data while preserving the class label \cite{AnabyTavor2019NotED,yang-etal-2019-exploring-pre}; (6) Distantly supervision and self-supervision try to introduce new training data from unlabeled text \cite{chen-etal-2017-automatically,ruiter-etal-2019-self}. As aforementioned, these frameworks cannot directly produce new suitable task-related examples for ECI. However, (1), (3), and (4) cannot guarantee the causality and well-formedness of new examples for ECI. Additionally, (2) and (5) are not easy to directly use external knowledge bases to generalize the event-related causal commonsense. Furthermore, (6) needs to design proprietary processing methods to generate ECI task-related training data. \citet{zuo-etal-2020-knowdis} solved the data lacking problem of ECI with the distantly supervised labeled training data. However, including the distant supervision, most of the existing text data augmentation methods for NLP tasks are task-independent frameworks. Therefore, we introduce a new learnable framework for augmenting task-related training data for ECI via dual learning enhanced with external knowledge. 
	
	\section{Generation with ConceptNet}
	To make a fair comparison, we introduce causal-related events from ConceptNet based on causal-related concepts, and obtain the causal sentence via the method in KonwDis \cite{zuo-etal-2020-knowdis} to further re-train MM \cite{ijcai2020-499}. Specifically, firstly, we obtain triples based on cause-related semantic relations from ConceptNet, such as \emph{Causes}, \emph{HasSubevent}, \emph{HasFirstSubevent}, \emph{HasLastSubevent}, \emph{MotivatedByGoal}, and \emph{CausesDesire} relations. Secondly, we assemble any two events from obtained causal triples to generate causal event pairs set and filter them via the filter of KonwDis. Next, we employ filtered causal event pairs to collect preliminary noisy labeled sentences from external documents via the DistantAnnotator of KonwDis. Then, we use the CommonFilter of KnowDis assisted with causal commonsense knowledge to pick out labeled sentences that express causal semantics between events. Finally, the refined causal sentences are input into LearnDA to generated ECI-related dual augmented training data and further train the MM to obtain MM+$ConceptAug$.
	
	\section{Main Experimental Environments and Other Parameters Settings}
	
	\subsection{Experimental Environments} 
	We deploy all models on a server with 250GB of memory and 4 TITAN Xp GPUs. Specifically, the configuration environment of the server is ubuntu 16.04, and our framework mainly depends on python 3.6.0 and PyTorch 1.0.
	
	\subsection{Other Parameters Settings} 
	All the final hyper-parameters for evaluation are averaged after 3 independent tunings on the development set. Moreover, the whole dual learning framework which includes event causality identifier and knowledge guided sentence generator takes approximately 5 minutes per epoch when training. According to the early stop strategy, the training rounds for different folds are different, and it takes about 20-30 rounds.

\end{document}